\documentclass[10pt, conference]{ieeeconf}
\usepackage[utf8]{inputenc}
\usepackage{cite}
  \usepackage[dvips]{graphicx}
  \usepackage{color}
  \usepackage{psfrag}

  \graphicspath{ {fig/} } % define default path for figures
\usepackage{amsmath}
\usepackage{amssymb}
\usepackage{bm}
\usepackage{algpseudocode}
  \usepackage[caption=false,font=footnotesize]{subfig}
\usepackage[hidelinks]{hyperref}

\usepackage{changes}

\usepackage[]{algorithm2e}

\usepackage{arydshln}

\IEEEoverridecommandlockouts % This command is only needed if you want to use the \thanks command
\title{\LARGE \bf
  Split Deep Q-Learning for Robust Object Singulation$^*$
}
\author{%
Iason Sarantopoulos$^{\dagger,1}$, Marios Kiatos$^{\dagger,1,2}$, Zoe Doulgeri$^{1}$ and Sotiris Malassiotis$^{2}$% <-this % stops a space
\thanks{$^*$The research leading to these results has received funding from the European Community’s Framework Programme Horizon 2020 under grant agreement No 871704, project BACCHUS.}
\thanks{$^\dagger$Authors have contributed equally.}
\thanks{$^1$Department of Electrical and Computer Engineering, Aristotle University of Thessaloniki, Thessaloniki, 54124 Greece {\tt\small \{iasons@, mkiatos@, doulgeri@eng.\}auth.gr}}%
\thanks{$^2$Information Technologies Institute (ITI) Center of Research and Technology Hellas (CERTH) 57001 Thessaloniki, Greece {\tt\small \{kiatosm,malasiot\}@iti.gr}}%
}

\begin{document}

\maketitle
\thispagestyle{empty}
\pagestyle{empty}
\begin{abstract}
Extracting a known target object from a pile of other objects in a cluttered environment is a challenging robotic manipulation task encountered in many robotic applications. In such conditions, the target object touches or is covered by adjacent obstacle objects, thus rendering traditional grasping techniques ineffective. In this paper, we propose a pushing policy aiming at singulating the target object from its surrounding clutter, by means of lateral pushing movements of both the neighboring objects and the target object until sufficient 'grasping room' has been achieved. To achieve the above goal we employ reinforcement learning and particularly Deep Q-learning (DQN) to learn optimal push policies by trial and error. A novel Split DQN is proposed to improve the learning rate and increase the modularity of the algorithm. Experiments show that although learning is performed in a simulated environment the transfer of learned policies to a real environment is effective thanks to robust feature selection. Finally, we demonstrate that the modularity of the algorithm allows the addition of extra primitives without retraining the model from scratch.
\end{abstract}

\section{Introduction}

Autonomous robots that can assist humans by performing everyday tasks have been a long standing vision of robotics and artificial intelligence. Robust robotic grasping of unknown objects in unstructured environments is an important skill for allowing service robots to perform within typical human environments, as well as for increasing the utility of industrial robots even further. However, due to the difficulty of robotic grasping, most of the existing solutions assume a collision free space around the target object \cite{bohg14, lenz15}. Other works \cite{eppner15a, sarantopoulos18a}, despite that they take into account the support surface by utilizing compliant contact with it during grasping, they still require the absence of surrounding clutter. Learning methods, \cite{morrison18, pinto16} have demonstrated increased success rates in grasping objects in cases there is sufficient free space for achieving a prehensile grasp. In cluttered scenes that do not allow prehension, manipulation techniques, aiming at singulation of a target object from its surrounding objects, have been proposed using a variety of actions, such as pushing, poking or picking and removing the obstacles. Singulation allows the existing grasping techniques to be applied on heavily cluttered scenes and, hence, increase the utility of the majority of grippers or multi-finger hands, as opposed to specialized end-effectors, like suction cups, used for non-prehensile target grasping \cite{mahler19}.

Existing singulation methods mainly use learning to achieve generalization of the skill in a range of complex environments. In contrast to hand-crafted algorithms, learning techniques can generalize across a wide variety of different situations by extracting the required statistical regularities from visual representations of the scenes. However, this comes at the cost of increased sample complexity, with general purpose learning algorithms requiring a large number of data in order to learn something useful. In this work, we focus on reducing the sample complexity for the singulation task, by modifying the architecture of the network representing the $Q$-function in a reinforcement learning scheme.

\begin{figure}[t!]
  \centering
  \includegraphics[width=0.45\textwidth]{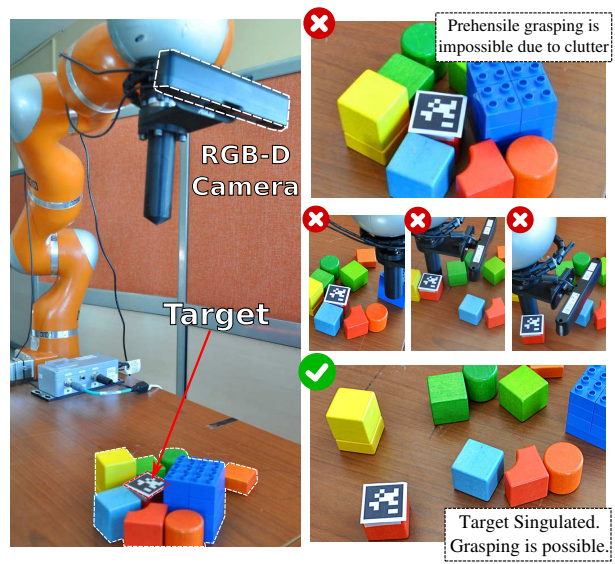}
  \caption{Typical scenario for singulating a target object.}
  \label{fig:task}
\end{figure}

In particular, in this paper, we use deep $Q$-learning \cite{mnih15} for learning a policy which robustly singulates a target object from its surrounding clutter, in order to enable a prehensile grasp (Fig. \ref{fig:task}). We learn a policy which uses pushing primitives as actions and we improve previous work \cite{kiatos19} by:
\begin{itemize}
  \item Splitting the $Q$-network so that the $Q$-function for each primitive action is learned independently. We demonstrate that splitting the $Q$-network results in faster convergence and increased success rate of the final policy for the object singulation task.
  \item Increasing the modularity of the algorithm, allowing the addition of an extra pushing primitive  without retraining the model from scratch.
  \item Training in a complex environment by making the possible scenes more random in terms of the number of obstacles, their dimensions and their pose around the obstacles.
  \item Showing that the learned policy can be effectively transferred to a real world setup thanks to robust feature selection.
\end{itemize}

In the following section the related work is presented. Section \ref{env} and \ref{mdp} describe the environment and the problem formulation, respectively. The proposed architecture is presented in Section \ref{learning}, its experimental evaluation using two pushing primitives in Section \ref{results} and  the benefit of its modularity, upon adding an extra pushing primitive, in Section \ref{modularity}. Finally, conclusions are drawn in Section \ref{conclusions}.

\section{Related Work}\label{related_work}

Pushing is a widely used action in robotic manipulation tasks. Lynch and Mason \cite{lynch1996stable} pioneered research on analytic models of push mechanics. Dogar \textit{et al.} \cite{dogar2010push} proposed a planning framework for push-grasping to reduce grasp uncertainty in a cluttered scene. However, they used known 3D models of objects and estimated the pose of each object in the scene. Without the prior knowledge of the objects, it is difficult to estimate the pose and physical properties of the objects, which can affect the efficiency of pushing actions.

Interactive-segmentation methods solve singulation tasks through a sequence of interactions given the segmentation of the scene \cite{hermans2012guided}, \cite{chang2012interactive}. Eitel \textit{et al.} \cite{eitel2017learning} trained a convolutional neural network on segmented images in order to singulate every object in the scene. Although, they removed the need for hand-engineered features they evaluated a large number of potential pushes in order to find the one with the highest probability. In  \cite{danielczuk2018linear}, the authors proposed two novel hand-crafted pushing policies for singulating objects in a bin with only one push using the euclidean clustering method for segmenting the scene into potential objects. Unfortunately, their results lacked any post-push grasp success rates. Danielczuk and Kurenkov \cite{danielczuk19} proposed a perception and decision system to pick from a bin a known target occluded by clutter. Specifically, they segment the scene using a variant of Mask R-CNN \cite{he2017mask} trained on synthetic depth images and feed each segment to an action selector which consists of different action policies. Then, the action selector determines which object to manipulate and the action with the highest quality metric is executed. However, the  approach greedily searches each action policy to determine the best action.
All the above works involve scene segmentation.

Recently, a lot of researchers cast this decision making problem in a reinforcement learning framework. Boularias \textit{et al.}\cite{boularias2015learning} explore the use of reinforcement learning for training control policies to select among push and grasp primitives. Given a depth image of the scene, they segment the scene, compute hand-crafted features for each action and execute the one with the highest probability. However, the agent learns to singulate two specific objects and must be retrained for a new set of objects. Zeng \textit{et al.}  \cite{zeng18} used $Q$-learning to train end-to-end two fully convolutional networks for learning synergies between pushing and grasping, which led to higher grasp success rates. Similar to \cite{zeng18}, our proposed method does not require scene segmentation but instead of end-to-end learning, we select a visual feature for training our network in order to increase  transferability of our trained policy from simulation to a real setup. Furthermore, our focus is not on clearing the scene from every object, thus, approaches such as \cite{zeng18} would result in unnecessary actions. Finally, Kiatos \textit{et al.} \cite{kiatos19} trained a deep $Q$-network to select push actions in order to singulate a target object from its surrounding clutter with the minimum number of pushes using depth features to approximate the topography of the scene. Although they demonstrated high singulation success rates in simulation, the network converged slowly, the singulation success dropped in the real world scenarios and the method included restricting assumptions for the height of the obstacles. We build upon these results by demonstrating faster convergence to the optimal policy and generalization to more scenes. Moreover, we demonstrate the modularity of the proposed approach by adding new action primitives.

\section{Environment and Assumptions}\label{env}
\begin{figure}[t]
  \centering
  \includegraphics[width=0.38\textwidth]{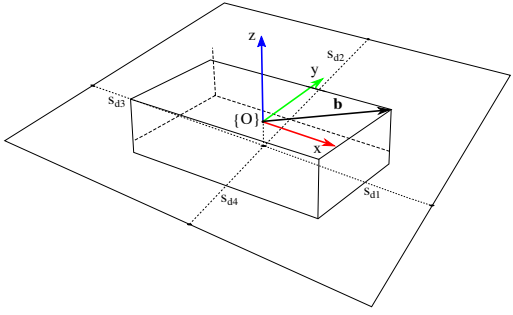}
  \caption{The representation of the target object and the support surface.}
  \label{fig:target}
\end{figure}

We consider the problem of singulating a target object of known position in a cluttered environment to assist grasping. More specifically, the typical environment for this task consists of a planar support surface which supports one target object and multiple obstacles adjacent to the target, similar to Fig. \ref{fig:task}. In this work we make the following assuptions:

\begin{itemize}
  \item A robotic system is available consisting of a fingertip which can be actuated on the Cartesian space and a depth camera which is able to capture depth information from a top view.
    \item Collision between the fingertip and an object can be detected. In a real setup this can be realized using force measurements.
  \item The frame placed on the geometric center of the target object, $\{O\}$, is known w.r.t. the camera frame at every timestep. In a real setup, this can be ensured by using an object detector, while in simulation the pose of the target can be obtained directly from the environment. Furthermore, the bounding box of the target object $\bm{b} = [b_1, \; b_2, \; b_3]^T$ is also known (Fig. \ref{fig:target}) and expressed in $\{O\}$.
   \item The pose, dimensions and the number of obstacles are random. In contrast to \cite{kiatos19}, our environment includes complex scenes in which all the objects can be of similar heights.
    \item The workspace of the scene is a rectangular with predefined dimensions placed on the support surface, with $\bm{s}_d \in \mathbb{R}^4$ denoting the 4 distances of the target object from the support surface's limits (Fig. \ref{fig:target}).
    \item Pushing actions are assumed to result in a 2D motion of the target object on the supported surface i.e. no flipping is expected during an episode.
\end{itemize}

The objective of this task is to singulate the target object from its surrounding obstacles with the minimum number of pushes in order to facilitate its proper grasping, while avoiding to throw the target off the support surface's limits. \emph{Singulation} means that the target object is separated from the closest obstacle by a minimum distance $d_{sing}$.

\section{MDP Formulation}\label{mdp}
We formulate the problem as an episodic Markov Decision Process (MDP) with time horizon $T$. An MDP is a tuple ($\mathcal{S}, \mathcal{A}, \mathcal{R}, \mathcal{T}, \gamma$) of continuous states $\mathcal{S}$, discrete actions $\mathcal{A}$, rewards $\mathcal{R}$, an unknown transition function $\mathcal{T}$ and a discount factor $\gamma$. With $t \in \{0, 1, \dots, T-1\}$ we denote the discrete timestep within an episode. The discrete-time dynamics of the system are $\bm{x}_{t+1} = \mathcal{T}(\bm{x}_t, u_t)$, with $\bm{x}_t \in \mathcal{S}$, $u_t \in \mathcal{A}$ and $\bm{x}_0$ given. In each timestep, the agent selects an action according to a policy $u_t = \pi(\bm{x}_t)$, observes a new state $\bm{x}_{t+1}$ and receives a reward $r(\bm{x}_t, u_t) \in \mathcal{R}$. Given the MDP the goal is to find an optimal policy $\pi^*$ that maximizes the total expected reward $G = \sum_{k=0}^{T}\gamma^{k}r(\bm{x}_{k}, u_{k})$. We use a version of the $Q$-learning algorithm \cite{watkins1989learning} to learn a policy $\pi(\bm{x}_t)$ that chooses actions by maximizing the action-value function i.e. the $Q$-function, $Q_{\pi}(\bm{x}_t, u_t) = \sum_{k=t}^{T}\gamma^{k-t}r(\bm{x}_{k}, u_{k})$, which measures the expected reward of taking action $u_t$ in state $\bm{x}_t$ at timestep $t$.

\begin{figure}[!t]
  \centering
  \subfloat[Pushing the target ($\bm{p}_0$ at the side of the target object, $\theta = \pi$).]{\includegraphics[width=0.45\textwidth]{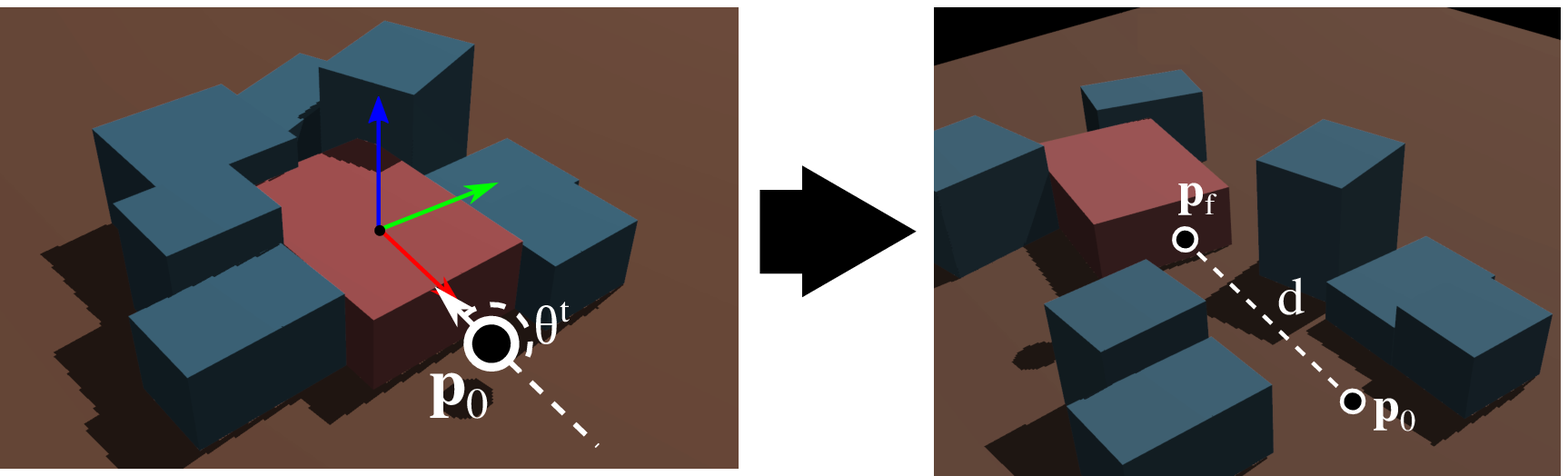}
  \label{fig:push_a}}
  \hfil
  \subfloat[Pushing an obstacle ($\bm{p}_0$ above the target object, $\theta = \frac{\pi}{4}$).]{\includegraphics[width=0.45\textwidth]{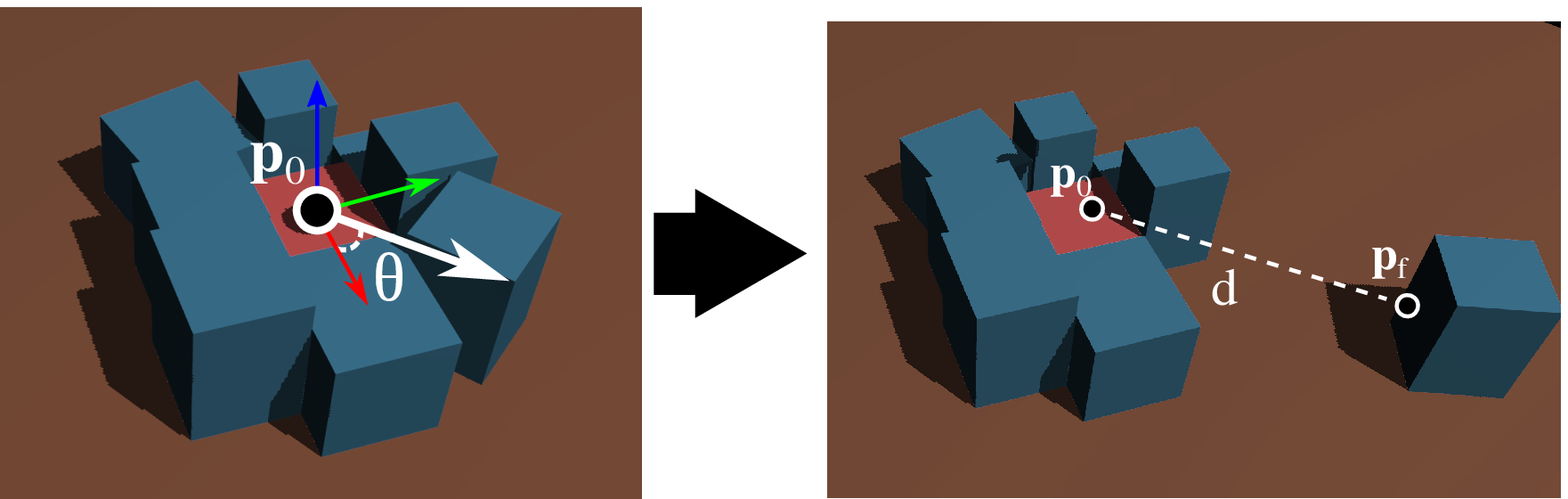}%
  \label{fig:push_b}}
  \hfil
  \caption{Illustration of the two pushing primitives before (left) and after (right) their execution. The initial point $\bm{p}_0$, pushing distance $d$ and angle $\theta$ are shown.}
  \label{fig:push}
\end{figure}

\subsection{Actions} \label{actions}
The robot can change the pose configuration of the objects by pushing them. In particular, we represent a pushing action, $\mathcal{P}$, similarly to \cite{kiatos19}, using the following tuple: $\mathcal{P} = (\bm{p}_0, \quad d, \quad \theta)$, where $\bm{p}_0 \in \mathbb{R}^3$ is the initial point that the push starts w.r.t. the object frame $\{O\}$, $d \in \mathbb{R}$ a predetermined pushing distance and $\theta \in [0, 2\pi)$ the angle defining the direction of the pushing action w.r.t. the $x$-axis of $\{O\}$. The motion of the push is parallel to the support surface and can be performed by generating a trajectory between $\bm{p}_0$ and $\bm{p}_f = [d \cos(\theta), \; d \sin(\theta), \; 0]^T$ expressed in $\{O\}$. In this work, we use two pushing primitives: 'Push target object' ($\mathcal{P}_{\tau}$) and 'push obstacle' ($\mathcal{P}_o$), which are shown in Fig. \ref{fig:push} and formally described as:
\begin{equation*}
  \mathcal{P}_\tau = \left(\begin{bmatrix} - (\sqrt{b_1^2 + b_2^2} + \epsilon) \cos(\theta) \\ - (\sqrt{b_1^2 + b_2^2} + \epsilon) \sin(\theta) \\ 0\end{bmatrix}, \quad d, \quad \theta \right)
\end{equation*}
\begin{equation*}
  \mathcal{P}_o = ([0, \;\; 0, \;\; b_3 + \epsilon]^T, \quad d, \quad \theta)
\end{equation*}
with $\epsilon$ a small offset used for compensating pose estimation errors, if any. To reduce the action space,  we discretize $\theta$ in $w$ discrete angles, $\theta_i = \frac{2\pi}{w}i, i = 0, \dots, w - 1$, resulting in $w$ different pushing directions for each of the two pushing primitives. Hence, the discrete action space of the MDP consists of $2w$ total actions, $u_t \in \mathcal{A} = \{0, 1, \dots, 2w - 1\}$, with the first $w$ actions corresponding to the primitive $\mathcal{P}_\tau$ and the last $w$ actions to $\mathcal{P}_o$.

\subsection{States}\label{states}
\begin{figure*}[t!]
	\centering
	\includegraphics[width=1\textwidth]{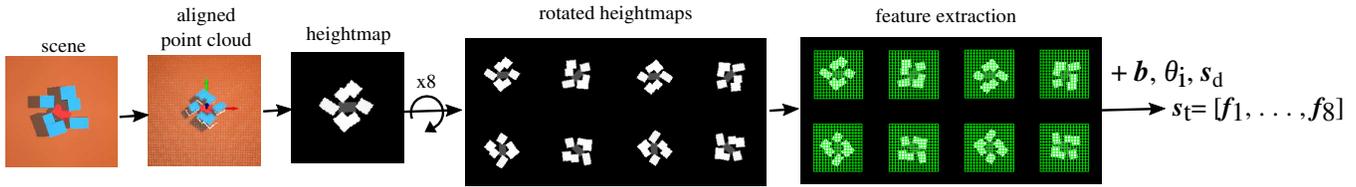}
	\caption{The pipeline of extracting the feature vector which represents the state.}
	\label{fig:state_pipeline}
\end{figure*}

We represent the state by a feature vector that describes the topography of the scene, with the pipeline for its extraction illustrated in Fig. \ref{fig:state_pipeline}. At each time step $t$ we estimate the pose of the target and we transform the acquired point cloud w.r.t. the target frame $\{O\}$. Then, we generate a heightmap $H$ where each $0.025 \times 0.025 \textrm{ cm}^{2}$ cell contains the highest z-value of points with corresponding $x$ and $y$ values. In contrast to \cite{kiatos19} we rotate the heightmap $H$ into $w$ orientations resulting to $w$ rotated heightmaps $h_i, i=0,\dots,w-1$ in order to simplify learning.
At each rotated heightmap $h_i$, we define a rectangular region of size $16 \times 16$ as shown in Fig. \ref{fig:state_pipeline}. For the $j$-th cell in this area we compute a feature as the average of the heightmap's values:

\begin{equation} \label{eq:feature calculation}
	z_{ij} = \frac{1}{c_x \cdot c_y} \sum_{x=x_1}^{x_2} \sum_{y=y_1}^{y_2} h_i(x, y)
\end{equation}
where $j=0,\dots,255$ and $c_x$ and $c_y$ are the dimensions of each cell. By concatenating the features from all regions we end up with the feature vector $\bm{z}_i = [z_{i0}, \dots, z_{i255} ]^T$.

To explicitly denote which cells belong to the target object, we add to each feature vector $\bm{z}_i$ the dimensions of the target's bounding box $\bm{b}$ and the rotation angle $\theta_i$ of each heightmap. Finally, we add to each feature vector the distances $\bm{s}_d$ of the target object from the support surface's limits, in order to avoid pushing off the limits of the support surface. Note that we rescale each feature vector $\bm{z}_i$, the bounding box $\bm{b}$ and the distances $\bm{s}_d$ at the range $[0, 1]$ using their respective maximum values, so that the final feature vectors are $\bm{f}_i = [\bm{z}_i, b_1, b_2, \theta_i, \bm{s}_d] \in \mathbb{R}^{263}$. The state $\bm{x}_t$ is then represented by $\bm{x}_t = [\bm{f}_0, \dots, \bm{f}_{w-1}]$.

\subsection{Rewards}
We define a sparse reward function. If the push results to the target's singulation, the reward is $r=+10$. If the push results to the target falling off the support surface the reward is $r=-10$. Furthermore, we punish any unintentional collision of the fingertip with some object by $r=-10$. Unintentional collision is considered any collision before reaching $\bm{p}_0$ (e.g. if the "pushing the target" primitive is attempted at $\bm{p}_0$ occupied by an obstacle). In any other case the reward is defined based on whether the push changes the scene or not. For faster convergence we punish 'empty' pushes (i.e. finger motion that does not result in object movement) as $r=-5$. If the push changes the scene we assign $r=-1$, instead of $0$ so that to penalize the total number of pushing actions.

\subsection{Terminal states}\label{terminal states}
The episode is terminated when the target object is singulated, it has fallen out of the support surface, an unintentional collision between the fingertip and some object is detected or the predefined maximum number of timesteps $T_{max}$ has been reached. The episode is considered successful only on the first case of target singulation. In any other case the episode is considered as failed.

\section{Split DQN}\label{learning}

\begin{figure}[!t]
  \centering
  \subfloat[Vanilla DQN]{\includegraphics[width=0.20\textwidth]{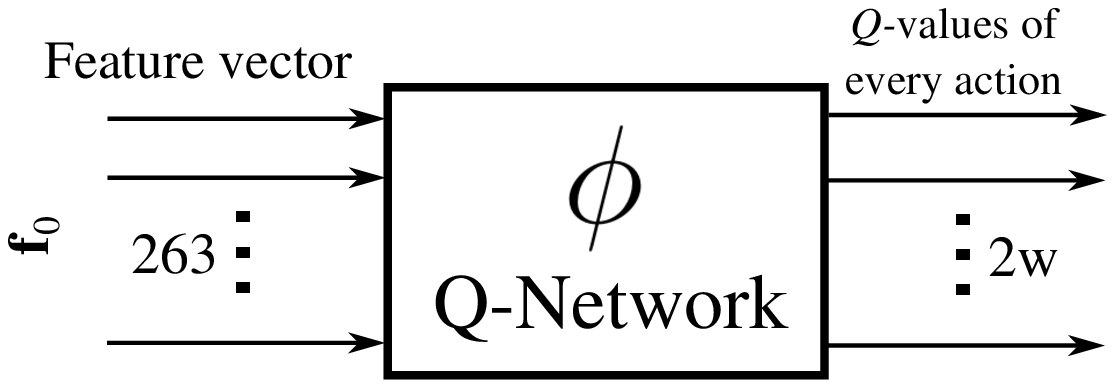}
  \label{fig:dqn_a}}
  \hfil
  \subfloat[Split DQN]{\includegraphics[width=0.26\textwidth]{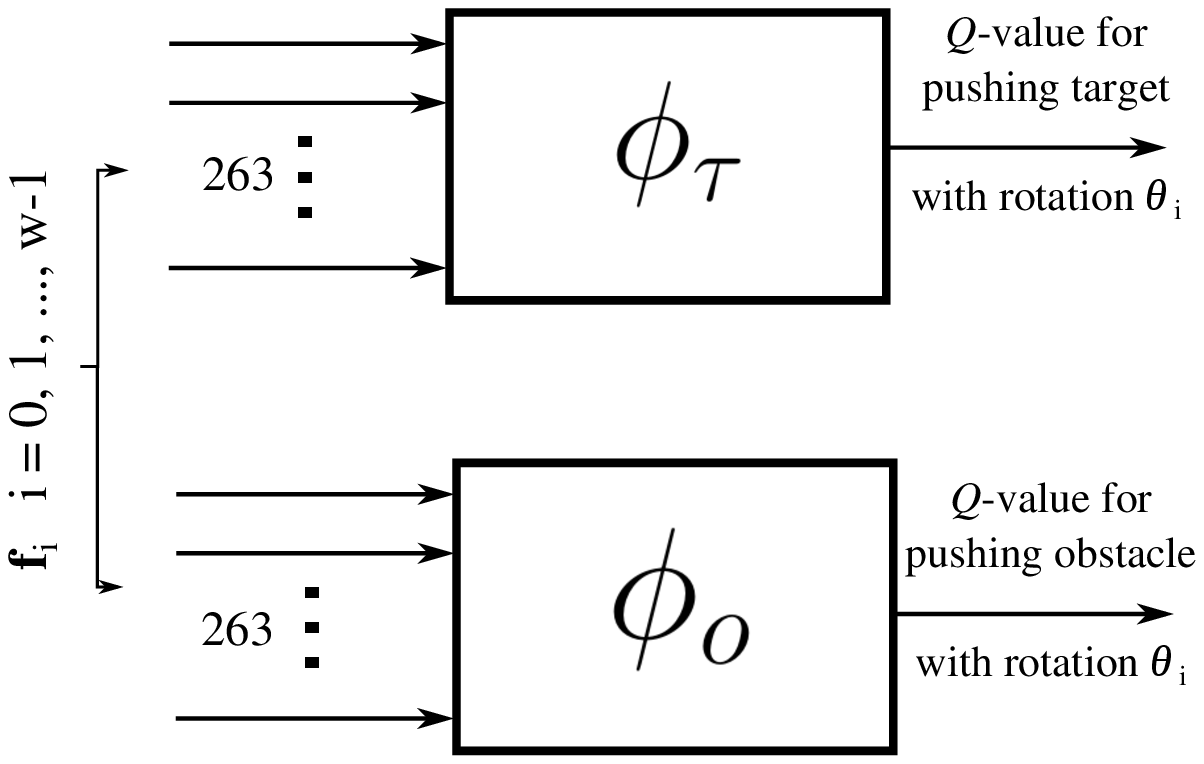}%
  \label{fig:dqn_b}}
  \hfil
  \caption{Singulation policy. (a) The vanilla DQN (b) The proposed split DQN architecture which uses different networks for each primitive action with the proposed rotation-invariant features fed into each network.}
  \label{fig:dqn}
\end{figure}

\subsection{Architecture}
Inspired by \cite{zeng18}, we modify the vanilla DQN \cite{mnih15} (Fig. \ref{fig:dqn_a}) by modelling our $Q$-function with two different networks $\phi_{\tau}$ and $\phi_o$ as shown in Fig. \ref{fig:dqn_b}. Each network is a feed forward fully connected network and corresponds to each of the pushing primitives, "push target" $\mathcal{P}_\tau$ and "push obstacle" $\mathcal{P}_o$. Both networks take as input the rotation-invariant feature vectors $\bm{f}_i$ and output a $Q$-value. The output of each network represents the future expected reward of executing the high level action $\mathcal{P}$ in orientation $\theta_i$. Hence, the $Q$-function in our case is represented as:
\begin{equation*}
Q (\bm{x}_t, u_t) =
\begin{cases}
\phi_\tau(\bm{f}_{u_t}), & \text{if } u_t \in \mathcal{A}_\tau = \{0, \dots, w - 1\} \\
\phi_o(\bm{f}_{u_t - w}), & \text{if } u_t \in \mathcal{A}_o = \{w, \dots, 2w-1\} \\
\end{cases}
\end{equation*}

This architecture exploits the proposed rotation-invariant features $\bm{f}_i$ (Section \ref{states}) to simplify learning the $Q$-function for pushing in different orientations, which means that we account only for horizontal pushes. Hence, the maximum $Q$-value is given by:
\begin{equation*}
\max_{u_t \in \mathcal{A}}Q(\bm{x}_t, u_t) = \max \left(\max_{u_t \in \mathcal{A}_{\tau}}{\phi_\tau(\bm{f}_{u_t})}, \max_{u_t \in \mathcal{A}_o}\phi_o(\bm{f}_{u_t - w})\right)
\end{equation*}

The intuition behind splitting the vanilla DQN is that each pushing primitive is correlated with features from different distributions. In particular, consider the different effect that the two pushing primitives have on the environment. On one hand, the "pushing target" action is effectively displacing multiple obstacles w.r.t. $\{O\}$ by usually removing the target object from a cluster of obstacles (see Fig. \ref{fig:push_a}). On the other hand, the "pushing obstacle" primitive displace only a small number of obstacles w.r.t. $\{O\}$, the ones along the path of the pushing direction (see Fig. \ref{fig:push_b}).
Having one monolithic network for all $2w$ actions, as in vanilla DQN, ignores this insight and slows down learning by trying to learn their more complex joint distribution. On the contrary, having one network dedicated for each pushing primitive results to each network training on data that come from the same distribution, which leads to faster convergence to the true $Q$-value.

Another advantage of this architecture, which can facilitate faster learning, is its inherent modularity. Specifically, if we want to add an extra primitive action, we do not need to retrain the whole network from scratch as dictated by the monolithic nature of the vanilla DQN, but we can add an extra network corresponding to the new primitive action and use pre-trained networks for the existing primitives.

\iffalse
\begin{figure}[!t]
  \centering
  \subfloat[]{\includegraphics[width=0.40\textwidth]{dqn_rot_explain_a}
  \label{fig:dqn_rot_explain_a}}
  \hfil
  \subfloat[]{\includegraphics[width=0.40\textwidth]{dqn_rot_explain_b}%
  \label{fig:dqn_rot_explain_b}}
  \hfil
  \caption{}
  \label{fig:dqn_rot_explain}
\end{figure}
\fi
\subsection{Training}
We use two replay buffers (one per primitive). In each timestep, we perform an explorative action  $u_t$ and we store the transition to the replay buffer corresponding to $u_t$. Then, we sample a minibatch of $K$ stored transitions $(x_k, r_k, u_k, x_k^{next})$ from this replay buffer and we train the corresponding network by minimizing the mean-squared error loss function:
\begin{align*}
	L &= \frac{1}{K}\sum_{k = 1}^K(Q^{\omega}(\bm{x}_k, u_k) - y_{k})^2 \\
	\text{with: } y_k &= r_k + \gamma \max_{u \in \mathcal{A}} Q^{\omega^{-}}(\bm{x}_k^{next}, u)
\end{align*}
where $\omega$ are the parameters of the corresponding primitive network for this timestep and $\omega^-$ are the parameters of its target network. We use "soft" target updates, rather than directly copying the weights. The weights of the target networks are then updated by slowly tracking the learned network's weights $\omega^{-} \leftarrow \tau\omega + (1 - \tau)\omega^{-}$ with $\tau \ll 1$ \cite{lillicrap2015continuous}. Notice that by updating only  the network that corresponds to the explorative action, we can prevent overfitting of the network for which no new samples are added to its buffer during exploration.

\section{Experiments}\label{results}
We executed a series of experiments\footnote{A video can be found in: \url{https://youtu.be/ef1MKgVkN0E}} in simulation and in a real world scenario to test our proposed approach. The policies are trained in simulation and are evaluated both in the simulated and the real environment. The goals of the evaluation experiments are 1) to investigate whether splitting DQN can help the Q-Network converge faster to the optimal policy, 2) to evaluate the quality of the derived policy with respect to the objective of the task, i.e. the singulation success and 3) to demonstrate the robust transfer to a real world setup.

\subsection{Simulated Environment}\label{simulated_env}
We use the MuJoCo physics engine \cite{todorov12} to advance the simulation after each action. We approximate the objects as rectangulars of random dimensions and the robotic finger as a floating sphere of radius $0.5$ cm. The number of obstacles is between $5$ and $8$, with the smallest possible bounding box $[1, 1, 0.5]^T$ cm and the largest $[3, 3, 2]^T$ cm, which means that there is a small chance to spawn a scene with similar height objects. The point cloud for the feature extraction is acquired by rendering the scene. The episode is terminated after $T_{max} = 20$ timesteps, in case none of the terminal states described in Section \ref{terminal states} have been reached. All the dynamic parameters of the simulated environment are kept to their default values. For the actions, we use $w=8$ different directions for each pushing primitive resulting to $16$ total available actions. Finally, we use $d_{sing} = 3$ cm, as the minimum distance for considering the target object as singulated.

\subsection{Policies and Training}
We train and compare two policies: DQN and Split DQN. \textbf{DQN} is a pushing policy that uses the vanilla DQN, shown in Fig. \ref{fig:dqn_a}, with two hidden layers with 140 units for each. \textbf{Split DQN} is the proposed pushing policy based on the splitting of the Q-Network into two different networks, each one corresponding to a primitive action and also integrating the rotated features as inputs, as shown in Fig. \ref{fig:dqn_b}. Each primitive network consists of two hidden layers with 100 units each. All the networks are trained using the Adam optimizer with learning rate $0.001$. We use an $\epsilon$-greedy exploration policy with $\epsilon$ decaying exponentially from $0.9$ to $0.25$ over $20$k timesteps and discount factor $\gamma = 0.9$. The replay buffers are preloaded with $1$k transitions each (acquired by random exploration) in order to decrease training time  and we sample batches consisting of $K = 64$ samples.

\begin{figure}[!t]
  \centering
  \includegraphics[width=0.37\textwidth]{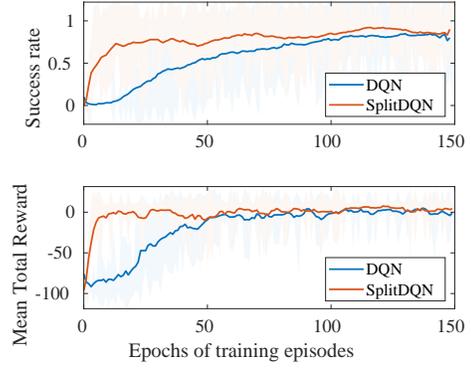}
  \caption{Training curves for the vanilla DQN and the proposed SplitDQN.}
  \label{fig:results}
\end{figure}

We train the above policies for $3000$ training episodes (resulting to approximately $10$k timesteps). Every $20$ training episodes (called here an epoch), we run $10$ testing episodes for estimating the progress of the success rate and the total reward of the policies. Fig. \ref{fig:results} shows the success rate and total reward per episode during training, averaged for each epoch. The results indicate that the proposed modifications led to faster convergence to the final policy compared to the vanilla DQN.

In order to evaluate the quality of the learned policies, we run $1000$ testing episodes of random generated scenes. For each case we evaluate the performance of the policies by measuring the success rate, the total number of pushes for the successful episodes and the expected reward. In order to put the results into perspective, we also use a random and a human policy in our comparison. \textbf{Random} selects a random action, according to a uniform distribution. \textbf{Human} is a policy in which a human selects one of the available $16$ actions, based on his own perceptual reasoning, by looking at the MuJoCo rendered scene. For practical reasons, we run $100$ testing episodes for the human policy. The results are demonstrated in Table \ref{tab:results} sorted by the success rate, showing that the proposed architecture of Split DQN not only converges faster to its final policy than DQN, but results to an improved policy for this environment presenting higher success rate, less number of pushes until singulation and higher mean reward than DQN.

\begin{table}[h!]
  \begin{center}
    \caption{Performance evaluation (Sorted by success rate)}
    \label{tab:results}
    \begin{tabular}{c | c c c c c}
      \hline
      \textbf{\scriptsize Policy} & \textbf{\scriptsize Success}  & \textbf{\scriptsize Mean} & \textbf{\scriptsize Std} & \textbf{\scriptsize Mean } & \textbf{\scriptsize Std }\\
                       & \textbf{\scriptsize rate} & \textbf{\scriptsize actions} &  \textbf{\scriptsize actions} & \textbf{\scriptsize reward} & \textbf{\scriptsize reward} \\
      \hline
      Human & 95.0\% & 2.46 & 0.88 & 7.51 & 4.36\\
      \textbf{SplitDQN} & \textbf{88.6\%} & \textbf{2.95} & {1.43} & \textbf{3.42} & 18.56 \\
      DQN & 77.1\% & 4.02 & 2.12 & -1.924 & 23.01\\
      Random & 22.1\% & 5.79 & 3.24 & -10.17 & 8.79\\
      \hdashline
      SplitDQN (Real) & 75.0\% & 2.71 & 1.18 & -1.37 & 5.60 \\
      \hline
    \end{tabular}
  \end{center}
\end{table}

\subsection{Policy Transfer in a real robotic system}
In this section, we evaluate the proposed policy in a real world scenario. Our real world setup consists of a 7DOF KUKA LWR4+ as the robotic arm with a wrist-mounted Xtion depth sensor (Fig. \ref{fig:task}). The object set consists of 20 objects, similar to the ones shown in Fig. \ref{fig:task}. We performed 40 experiments by placing the objects in random poses. The conducted experiments consist of the following steps. At first, the object detection algorithm runs on an acquired RGB-D image. Specifically, we use Apriltags \cite{olson2011apriltag} to find the pose of the target object by attaching one Apriltag on its top surface. Then, the robot chooses and executes a push action. The pushes are implemented as trajectories given to a Cartesian impedance controller, used for safety. This procedure is repeated until a terminal state (see Section \ref{terminal states}) is reached. Although the results indicate that the learned policy is robustly transferred to a real world scenario, a drop in success rate compared to simulation is observed. The difference in success rate is accounted to pose estimation errors of the target object, noisy data and discrepancy in the physics between simulation and real environment.

\section{Evaluating the architecture's modularity}\label{modularity}
In order to evaluate the modularity of the proposed architecture, we introduce an extra pushing primitive $\mathcal{P}_{extra}$. The initial position of this primitive is $\bm{p}_0 = [-\alpha\cos(\theta),\; -\alpha\sin(\theta),\; 0]^T$, with $\alpha = 25$ cm, the support surface's dimensions, which means that this primitive is similar to the ''push target'' primitive with the difference of starting the push outside the scene. Its advantage is that it can solve scenes like the one illustrated in Fig. \ref{fig:equal_height}, in which all the objects have similar heights and the existing two primitives cannot provide any solution. This primitive can approach from outside and break the cluster of objects. Its disadvantage is that the total pushing distance is large, which means that 1) the linear Cartesian trajectory required for the push might not be always realizable in the arm's joint space 2) increases the duration of the robotic action. For these reasons, we penalize the actions related to this primitive by $-5$.

\begin{figure}[!t]
  \centering
  \includegraphics[width=0.35\textwidth]{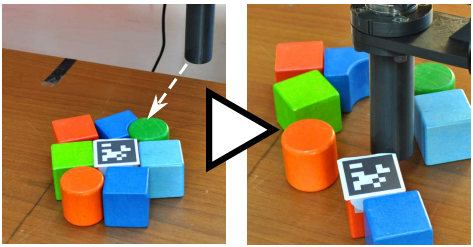}
  \label{fig:equal_height}
  \caption{A complex scene with objects of equal heights which requires the extra primitive of breaking the cluster.   %rendering the two standard pushing primitives inadequate to solve this scene. An extra primitive which approaches from outside the scene can break the cluster of objects, facilitating the use of the two standard primitives.
  }
  \label{fig:equal_height}
\end{figure}

\begin{figure}[!t]
  \centering
  \includegraphics[width=0.37\textwidth]{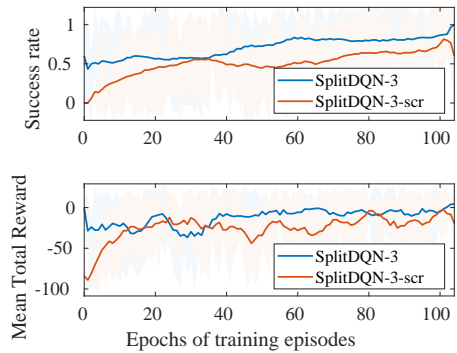}
  \caption{Training curves using an extra primitive  in a more complex environment with (SplitDQN-3) and  without (SplitDQN-3-scr) the use of pretrained networks.}
  \label{fig:results_n}
\end{figure}

We train the agent in a more complex environment than the one used in Section \ref{simulated_env} to demonstrate the utility of the extra primitive and the modularity of the proposed architecture. The difference is that we increase the probability to $20$\% of spawning scenes similar to the one in Fig. \ref{fig:equal_height} and we increase the maximum number of obstacles to $13$. First, we train with the two standard primitive actions (\textbf{SplitDQN-2}) and then we train with the extra primitive (\textbf{SplitDQN-3}) using 3 networks, with the first two using the pretrained networks of SplitDQN-2 and the third randomly initialized. Adding the extra primitive results in increased success rate ($83.4$\%), in contrast to using the first two primitives ($59.6$\%) (see Table \ref{tab:results_n}). Furthermore, we train the three networks from scratch (\textbf{SplitDQN-3-scr}). Although, SplitDQN-3 and SplitDQN-3-scr result to equally high success rates, SplitDQN-3 converges to the final policy faster (Fig. \ref{fig:results_n}), demonstrating the benefit of using pretrained primitive networks, which stems from the increased modularity of the architecture.

\begin{table}[h!]
  \begin{center}
    \caption{Performance evaluation for a more complex environment}
    \label{tab:results_n}
    \begin{tabular}{c | c c c c c}
      \hline
       \textbf{\scriptsize Policy} & \textbf{\scriptsize Success}  & \textbf{\scriptsize Mean} & \textbf{\scriptsize Std} & \textbf{\scriptsize Mean} & \textbf{\scriptsize Std}\\
                       & \textbf{\scriptsize rate} &  \textbf{\scriptsize actions}& \textbf{\scriptsize actions}  & \textbf{\scriptsize reward} & \textbf{\scriptsize reward} \\\hline
      SplitDQN-3 & {83.4\%} & {3.19} & {1.43} & {-2.64} & {20.92} \\
      %SplitDQN-ex & {69.8\%} & {3.48} & {1.58} & {-4.15} & {21.30} \\
      %SplitDQN-ex & {86.1\%} & {2.94} & {1.55} & {0.878} & {10.28} \\
      SplitDQN-2 & 59.6\% & {4.42} & {1.77} & {-20.35} & {40.95} \\
      \hline
    \end{tabular}
  \end{center}
\end{table}

\vspace{-0.6cm}\section{Conclusions and Future Work}\label{conclusions}
In this paper, we propose Split DQN, a variant of Deep Q-Network, for learning optimal push policies in order to singulate a target object from its surrounding clutter. We show that splitting the vanilla DQN to a set of subnetworks, one for each push primitive action, improved the convergence rate as well as the quality of the final policy and produced a modular architecture facilitating the addition of new action primitives. Results show that the learned policy can be robustly transferred to a real world scenario. In our future work we will investigate the method's performance with more primitive actions, e.g. pick and place, and the training of the robot in even more complex environments e.g. scenes in which the target object is covered by other objects from the top. Finally, we will explore whether continuous actions can produce optimal policies for this type of complex environments or not.

\bibliographystyle{IEEEtran}
\bibliography{references.bib}

\end{document}